\documentclass[sigconf]{acmart}

\usepackage{multirow}
\usepackage{pifont}
\usepackage{colortbl}
\definecolor{LightGreen}{RGB}{2, 125, 86}

\AtBeginDocument{%
  }

\copyrightyear{2026}
\acmYear{2026}
\setcopyright{cc}
\setcctype{by}
\acmConference[MM '26]{Proceedings of the 34th ACM International Conference on Multimedia}{November 10--14, 2026}{Rio de Janeiro, Brazil}
\acmBooktitle{Proceedings of the 34th ACM International Conference on Multimedia (MM '26), November 10--14, 2026, Rio de Janeiro, Brazil}
\acmDOI{10.1145/3767308.3836480}
\acmISBN{979-8-4007-2213-4/2026/11}
\begin{document}

\title{Beyond Pattern Matching in 3D MLLMs: Part-Level Functional Reasoning for Generalizable Recognition}

\author{Weihong Li}
\affiliation{%
  \institution{Tsinghua Shenzhen International Graduate School, Tsinghua University}
  \city{Shenzhen}
  \country{China}}
\email{li-wh24@mails.tsinghua.edu.cn}

\author{Hao Jiang}
\affiliation{%
  \institution{Tsinghua Shenzhen International Graduate School, Tsinghua University}
  \city{Shenzhen}
  \country{China}}
\email{h-jiang24@mails.tsinghua.edu.cn}

\author{Yaohua Zha}
\affiliation{%
  \institution{Tsinghua Shenzhen International Graduate School, Tsinghua University}
  \city{Shenzhen}
  \country{China}}
\email{zyh1614399882@gmail.com}

\author{Tao Dai}
\authornote{Corresponding author.}
\affiliation{%
  \institution{College of Computer Science and Software Engineering, Shenzhen University}
  \city{Shenzhen}
  \country{China}}
\email{daitao.edu@gmail.com}

\author{Shu-Tao Xia}
\affiliation{%
  \institution{Tsinghua Shenzhen International Graduate School, Tsinghua University}
  \city{Shenzhen}
  \country{China}}
\email{xiast@sz.tsinghua.edu.cn}

\begin{abstract}
3D MLLMs recognize objects via
pattern matching, which fails when geometrically similar objects
require functional reasoning to disambiguate (e.g., dresser
vs.\ wardrobe).  Our per-sample
\emph{Geometric Confusion Score} (GCS) exposes this failure: baseline
accuracy is 33.7\% on ModelNet40's most ambiguous 20\% versus 84.0\% on its
least ambiguous 20\%.  We identify a
\emph{function-as-annotation} pattern, in which functional language
decorates a known identity instead of supporting inference, and hypothesize
that reversing this information flow should concentrate improvements on
the most ambiguous samples.  PFIR
(Part-level Functional Inference for Recognition) reverses that flow.
Built on PartNeXt part segmentation, its data pipeline produces 507K QA
pairs with functional CoTs (geometry $\to$ function
$\to$ identity), and its progressive curriculum culminates in dual-objective
GRPO: strict functional reasoning and descriptive grounding for Objaverse's
diverse distribution.  The two RL objectives yield nearly orthogonal task
vectors ($\cos = 0.056$), enabling composition via Task Arithmetic.
Under a strict dual OOD protocol (Objaverse $\to$ ModelNet40;
XYZRGB $\to$ XYZ-only), PFIR raises zero-shot accuracy
from 60.5\% to 66.7\%, with gains concentrated
$2.6\times$ more on geometrically ambiguous samples---supporting
part-level functional reasoning as an effective route to more generalizable
3D recognition.
Code: \url{https://github.com/this233/pfir-3d}.
\end{abstract}

\ccsdesc[500]{Computing methodologies~Computer vision}
\ccsdesc[500]{Computing methodologies~Natural language processing}

\keywords{3D Multi-Modal LLM, Point Cloud Understanding, Functional Reasoning, Zero-Shot Classification, Reinforcement Learning}

\maketitle

\section{Introduction}

\begin{figure}[!t]
  \centering
  \begin{minipage}[t]{0.48\columnwidth}
    \centering
    \includegraphics[width=\linewidth,height=0.8\columnwidth,keepaspectratio]{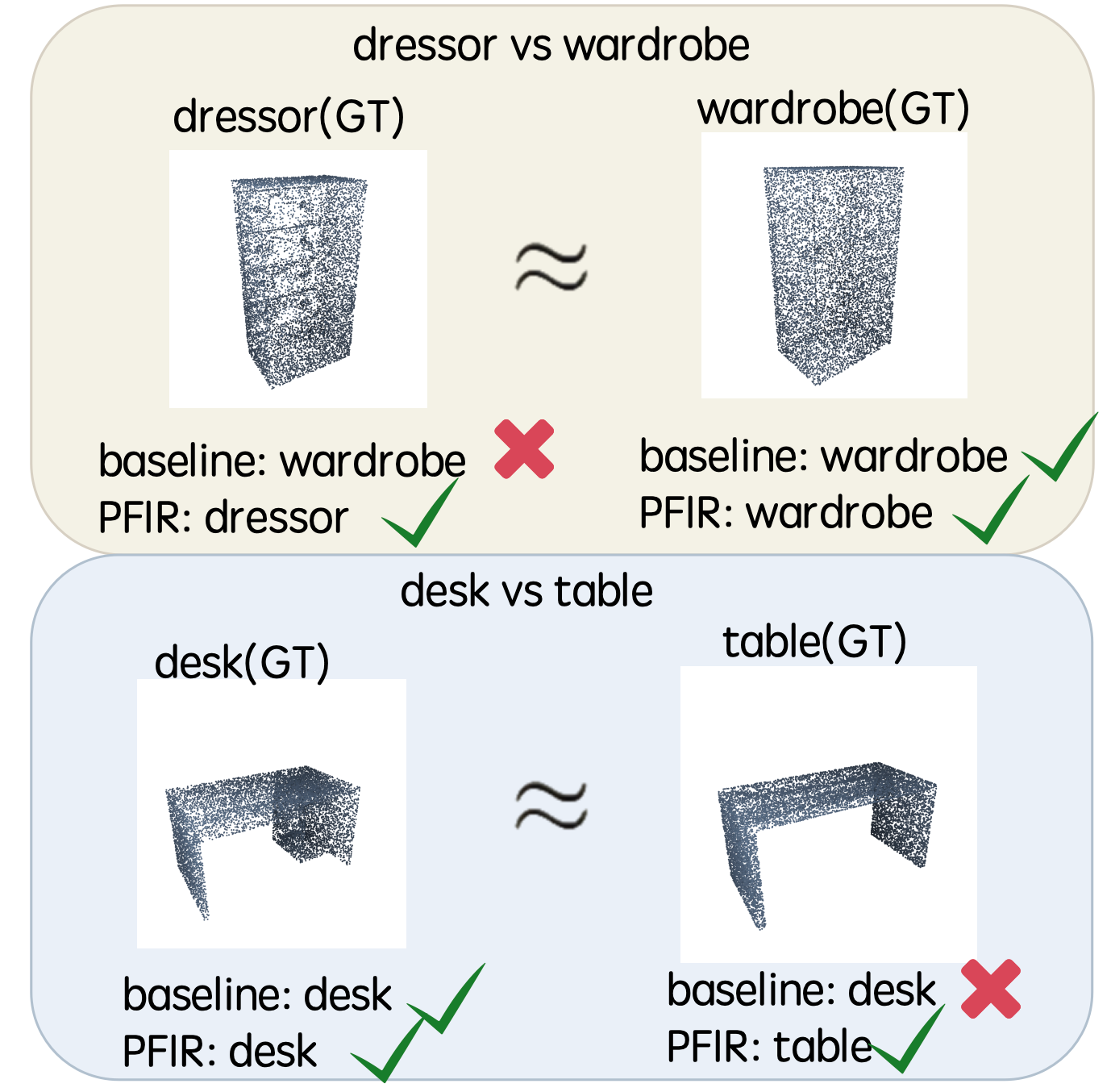}\\[0.2em]
    {\small\textbf{(a)} Geometrically ambiguous pairs}
  \end{minipage}
  \hfill
  \begin{minipage}[t]{0.50\columnwidth}
    \centering
    \includegraphics[width=\linewidth,height=0.8\columnwidth,keepaspectratio]{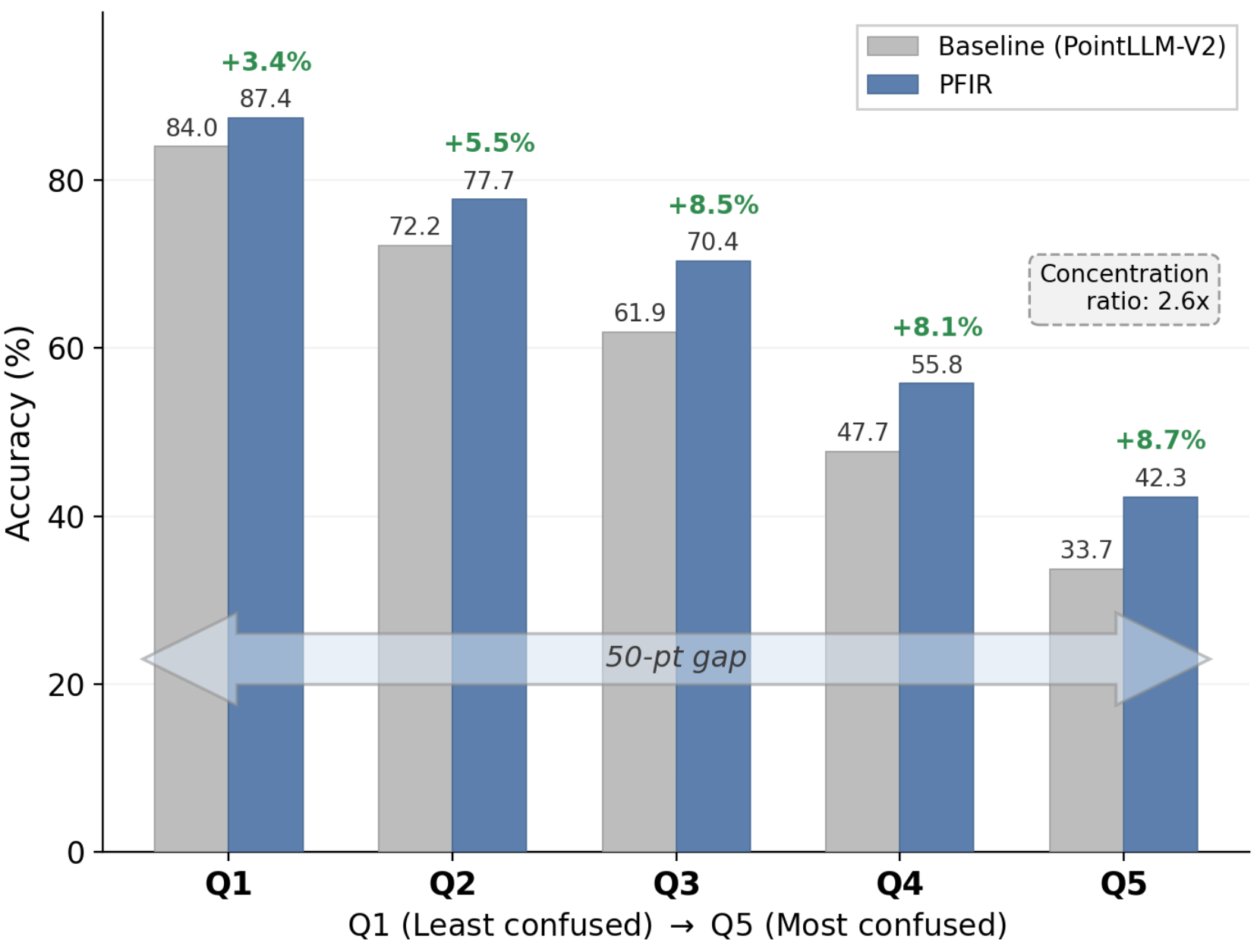}\\[0.2em]
    {\small\textbf{(b)} GCS stratified performance}
  \end{minipage}\\[0.4em]
  \includegraphics[width=\columnwidth]{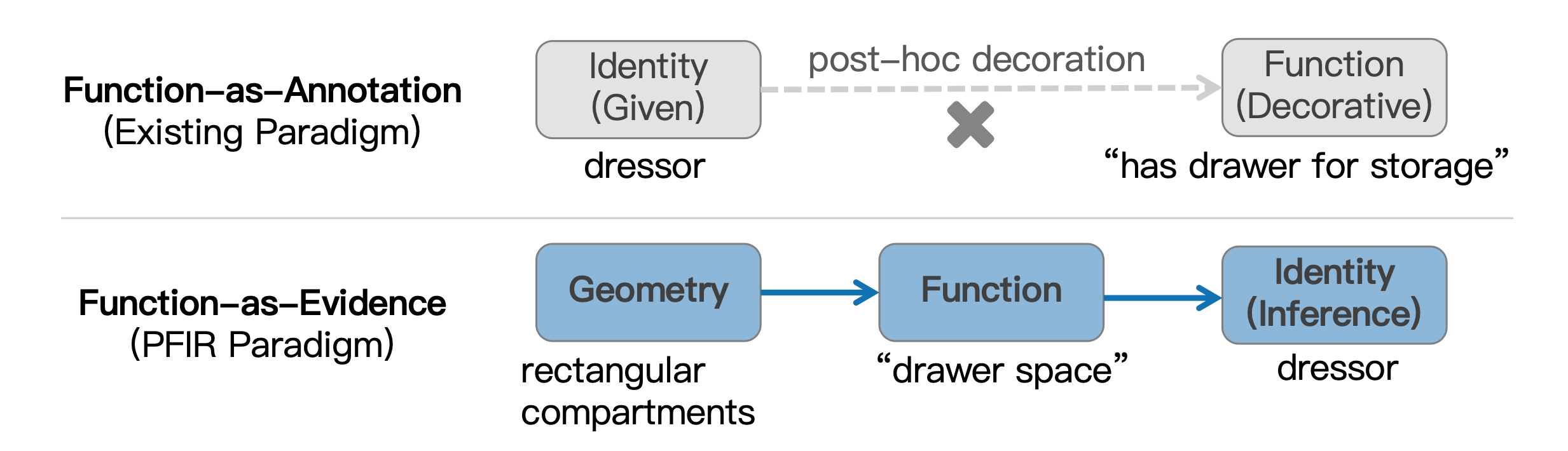}\\[0.2em]
  {\small\textbf{(c)} Function-as-Annotation vs.\ Function-as-Evidence}
  \caption{\textbf{Motivation.} Pattern matching collapses on geometrically
    ambiguous objects~\textbf{(a)}.  A per-sample GCS reveals that the
    hardest 20\% reach only 33.7\% baseline accuracy.  PFIR addresses this
    via part-level functional reasoning, with gains concentrated
    $2.6\times$ more on the most confused quintile~\textbf{(b)}, by shifting
    from function-as-annotation to function-as-\textbf{evidence}~\textbf{(c)}.}
  \Description{Teaser figure: confusion-prone 3D object pairs,
    GCS-stratified accuracy comparison, and information-flow contrast
    between function-as-annotation and function-as-evidence paradigms.}
  \label{fig:teaser}
\end{figure}

3D multimodal large language models (MLLMs) have advanced rapidly through
stronger encoders, larger datasets, and richer part-level
annotations~\cite{pointllm,pointllmv2,shapellm,gpt4point,lemon}.  Yet
their underlying recognition strategy has remained constant:
\emph{pattern matching}---comparing a point cloud against learned prototypes to find the
closest category.  This suffices when categories possess distinctive
silhouettes, as with the stylized 3D art dominating Objaverse, but collapses
when real-world taxonomies demand functional discrimination.  A dresser and
a wardrobe are both rectangular boxes; a sink and a bathtub are both concave
vessels; a desk and a table both have flat surfaces with legs.  To tell them
apart, one must reason about \emph{what the parts do}, not merely
\emph{what the whole looks like}---much as humans fall back on part-level
functional inference when global shape alone is
insufficient~\cite{biederman1987,gibson1979}.

We quantify this limitation with a per-sample \emph{Geometric Confusion
Score} (GCS), which measures feature-space proximity to the nearest
wrong-class centroid (Section~\ref{sec:problem}).  GCS stratification
reveals that baseline accuracy falls from 84.0\% on the least ambiguous
ModelNet40 quintile to 33.7\% on the most ambiguous, a 50-point gap hidden
by aggregate accuracy.

Although 27.4\% of training samples contain function-related
keywords, a contributing factor may be how information is organized
in Cap3D~\cite{cap3d} captions and PointLLM-V2's part descriptions.  Data
predominantly follow a \emph{function-as-annotation}
pattern: the object identity is stated first and functional language
decorates it post-hoc.  The resulting training signal therefore emphasizes
describing function \emph{given} identity rather than inferring identity
\emph{from} function.
We hypothesize that reversing this information flow---organizing
functional information as \emph{evidence} for identity inference
(geometry $\to$ function $\to$ identity)---should yield improvements
that concentrate disproportionately on the most geometrically ambiguous
samples, precisely where pattern matching fails.

We test this with \textbf{PFIR} (\textbf{P}art-level \textbf{F}unctional
\textbf{I}nference for \textbf{R}ecognition).  A PartNeXt-based
pipeline~\cite{partnext} produces 507K QA pairs whose chains follow the
function-as-evidence direction.  A three-stage curriculum---multi-source
alignment, chain-of-thought SFT, and GRPO~\cite{deepseekr1}---progressively
moves from part-aware perception and format imitation to active functional
discovery.

Because precise naming is unreliable for stylized and abstract Objaverse
objects, strict functional rewards may over-commit to incorrect identities.
We therefore pair functional reasoning with a descriptive-grounding reward
that favors rich geometric descriptions under uncertainty.  Their nearly
orthogonal task vectors ($\cos = 0.056$) support complementary composition
via Task Arithmetic~\cite{ilharco2023task}.  Under dual OOD shifts
(Objaverse $\to$ ModelNet40; XYZRGB $\to$ XYZ-only), PFIR raises
instruction-prompt accuracy from 60.5\% to 66.7\%, with gains $2.6\times$
more concentrated on the most ambiguous samples, supporting the hypothesis.

Our main contributions are:
\vspace{-1em}
\begin{itemize}
  \item A per-sample Geometric Confusion Score that quantifies the
    pattern-matching bottleneck and provides a prediction-independent
    diagnostic of geometric ambiguity.
  \item A part-level functional data pipeline (507K QA pairs with
    geometry $\to$ function $\to$ identity chains) and a three-stage
    curriculum from alignment through CoT SFT to dual-objective GRPO.
  \item Evidence that functional reasoning and descriptive grounding
    yield near-orthogonal task vectors, enabling complementary
    composition.  PFIR advances instruction-prompt OOD accuracy by +6.8
    percentage points , with gains
    $2.6\times$ concentrated on the most ambiguous quintile.
\end{itemize}

\section{Related Work}

\paragraph{3D Multimodal LLMs.}
PointLLM~\cite{pointllm} pioneered point cloud--language alignment and the
generative zero-shot paradigm: open-ended answers without candidate labels,
unlike discriminative methods~\cite{ulip2,openshape,uni3d}.  Later work
improves encoder quality (ShapeLLM~\cite{shapellm} via ReCon++), data scale
(GPT4Point's 1M pairs~\cite{gpt4point}), or architecture (MiniGPT-3D's 2D
priors~\cite{minigpt3d}; LEMON's unified tokenization~\cite{lemon}).
Complementary point-cloud research studies parameter-efficient transfer,
compact masked modeling, cross-domain self-supervision, and curvature
cues~\cite{zha2023instance,zha2025pma,zha2024towards,zha2024lcm,
zha2025point,zha2026casl}.  Our direct baseline, PointLLM-V2~\cite{pointllmv2},
improves perception granularity with SAM-based part segmentation~\cite{sam}
and 1.8M GPT-4o annotations~\cite{gpt4o}.  Its data, however, still describes
part functions given identity rather than training identity inference from
parts.

\paragraph{Part Understanding and Affordance.}
PartNet~\cite{partnet} and PartNeXt~\cite{partnext} provide hierarchical
part annotations, while Part-X-MLLM~\cite{partxmllm} unifies part-level
tasks through language.  3D-AffordanceLLM~\cite{3daffordancellm},
AffordBot~\cite{affordbot}, 3DAffordSplat~\cite{3daffordsplat}, and
GEAL~\cite{geal} make affordance the end-task; PFIR instead uses part
function as intermediate evidence along the geometry $\to$ function $\to$
identity chain.  PointCoT~\cite{pointcot} and SceneCOT~\cite{scenecot}
supervise geometry-grounded reasoning chains for 3D reasoning and question
answering, whereas PFIR's curriculum and RL objectives target the
recognition strategy itself.

\paragraph{RL for Multimodal Reasoning.}
GRPO~\cite{deepseekr1} extends to 2D
vision-language~\cite{r1vl,visionr1,visionaryr1,syngrpo} and 3D scene QA
(3D-R1~\cite{3dr1}), video-based 3D (3D-RFT~\cite{3drft}), and manipulation
(Embodied-R1~\cite{embodiedr1}).  To our knowledge, PFIR is its first use
for point-cloud \emph{object} understanding.  Rather than merely optimizing
answer quality, its RL stage activates the function-as-evidence strategy
learned during cold-start SFT for OOD generalization.  Dual objectives
further balance bold functional inference against descriptive grounding
across diverse assets.

\section{Method}

\begin{figure*}[!t]
  \centering
  \includegraphics[width=\textwidth]{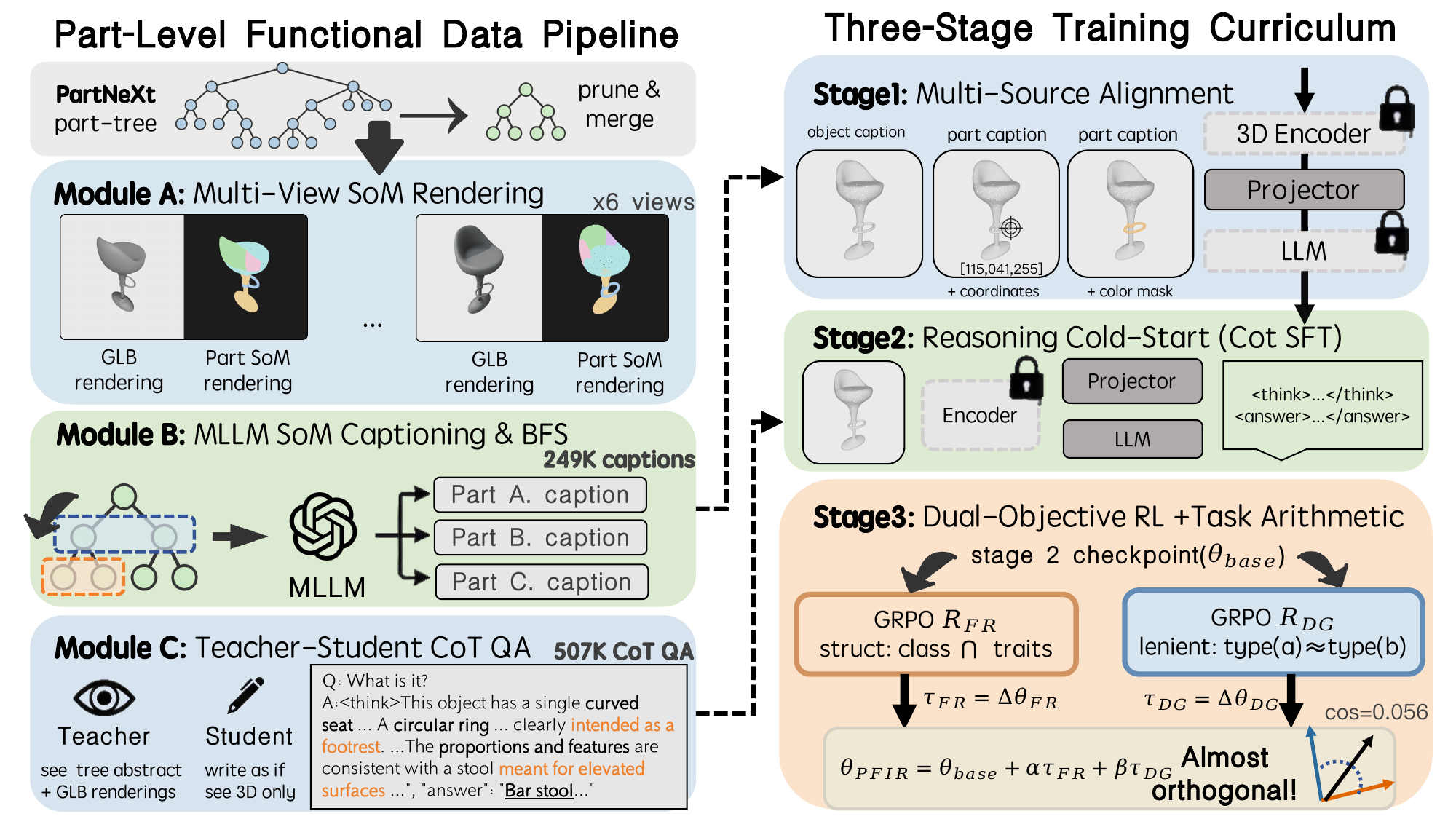}
  \caption{Overview of the proposed PFIR framework.  The part-level functional
    pipeline produces 507K QA pairs with function-as-evidence reasoning
    chains.  The three-stage curriculum progressively builds function-guided
    recognition: Stage~1 aligns part-aware representations, Stage~2 teaches
    the reasoning format via SFT, and Stage~3 trains two complementary RL
    experts (functional reasoning + descriptive grounding) composed via Task
    Arithmetic.}
  \Description{Overview of the PFIR framework: part-level functional data pipeline
    and three-stage training curriculum.}
  \label{fig:framework}
\end{figure*}

\subsection{Problem Formulation and Diagnostic Metric}
\label{sec:problem}

\paragraph{Limitations of pattern matching.}
Let $f$ denote a 3D MLLM that maps a point cloud $P$ to a textual prediction
$y = f(P)$.  Current models implicitly learn a pattern-matching function
that compares the representation of $P$ (holistically or part-by-part)
against learned category prototypes.  This suffices when categories possess
visually distinctive silhouettes but fails when multiple categories share
similar geometry yet differ fundamentally in functional identity.

\paragraph{Geometric Confusion Score (GCS).}
To quantify this limitation at the \emph{individual sample} level without
using any evaluated model's predictions, we define a per-sample Geometric
Confusion Score.  The score uses a fixed pretrained 3D encoder and is
therefore prediction-independent, rather than independent of all model
choices.
We extract shape features for every test sample and class centroids for
each of the $K$ classes in the evaluation set using the pretrained
Point-BERT encoder~\cite{pointbert}.  For a sample $P_i$ with
ground-truth class $c_i$, its GCS is the maximum cosine similarity to any
non-ground-truth class centroid:
\begin{equation}
  \mathrm{GCS}(P_i) = \max_{j \neq c_i}\;\cos\!\bigl(\mathbf{z}_{P_i},\;
  \boldsymbol{\mu}_j\bigr),
  \label{eq:gcs}
\end{equation}
where $\mathbf{z}_{P_i}$ is the feature of sample $P_i$ and
$\boldsymbol{\mu}_j$ is the centroid of class $j$.  A higher GCS indicates
greater geometric ambiguity---the sample's shape is close to at least one
wrong-class prototype.  We partition the evaluation set into quintiles
by GCS (Q1 = least confused, Q5 = most confused) and report per-quintile
accuracy.  This metric depends solely on the geometric structure of the
evaluation data and is independent of every evaluated model's predictions;
it is not used for training, model selection, or reward computation, while
providing finer-grained diagnostics than any class-level aggregate.

\paragraph{Function-as-annotation \emph{vs.}\ function-as-evidence.}
Even when existing training data contains functional vocabulary, the
information follows a \emph{function-as-annotation} pattern: the object
identity is stated first and functional descriptions serve as post-hoc
embellishments.  The model thus learns to \emph{decorate} known identities
with functional language, rather than to \emph{reason} from functional
observations toward identity.  We hypothesize that this misalignment may
contribute to the GCS-correlated accuracy degradation observed above: models
equipped with functional vocabulary may still struggle to leverage it for
disambiguation when the training signal does not require forward reasoning
from function to identity.
PFIR reorganizes functional information into
a \emph{function-as-evidence} pattern.  Concretely, for a point cloud $P$
with part set $\{p_k\}$, the desired reasoning chain factorizes identity
prediction as:
\begin{equation}
  p(y \mid P) = \sum_{\{f_k\}} p\!\bigl(y \mid \{f_k\}\bigr)\;
    \prod_{k} p\!\bigl(f_k \mid \mathrm{geom}(p_k)\bigr),
  \label{eq:chain}
\end{equation}
where $\mathrm{geom}(p_k)$ denotes the geometric observation of part $p_k$
  and $f_k$ is the inferred function.  PFIR seeks to infer identity from an
  \emph{ensemble of part functions} rather than relying solely on holistic
  shape matching.  Existing training data predominantly conditions
  functional descriptions on a known identity and therefore does not
  explicitly teach the forward reasoning direction.

\subsection{Overview}

With the problem and diagnostic metric established, we now present PFIR,
a three-stage framework that progressively builds function-guided
recognition capability in 3D MLLMs (Figure~\ref{fig:framework}).
We adopt the PointLLM-V2 architecture with its Point-BERT encoder and
replace the LLM backbone with Qwen3-VL-8B (same 8B parameter class as
the original LLaMA-3.1-8B; the Stage~1 control in Section~\ref{sec:ablation}
indicates no accuracy advantage from this swap alone).  PFIR
proceeds in three stages:
(1)~multi-source alignment that establishes part-aware perceptual
foundations (Section~\ref{sec:stage1}); (2)~chain-of-thought supervised
fine-tuning that teaches the format of function-as-evidence reasoning
(Section~\ref{sec:stage2}); and (3)~dual-objective reinforcement learning
with Task Arithmetic composition that transitions the model from format
imitation to active functional discovery (Section~\ref{sec:stage3}).
Stages~1 and~2 are powered by a unified part-level functional data pipeline
(Section~\ref{sec:pipeline}); Stage~3 trains on the re-annotated Objaverse
descriptions from Stage~1, requiring the model to autonomously apply the
reasoning strategy learned in Stage~2.

\subsection{Part-Level Functional Data Pipeline}
\label{sec:pipeline}

The sole input to our pipeline is PartNeXt's~\cite{partnext} \emph{coarse}
part-tree annotation---a hierarchy of nodes carrying rough part names---and
the corresponding per-part mesh segmentation.  From this minimal
supervision, the pipeline automatically produces structured triplets
$\{\text{question},\;\text{think},\;\text{answer}\}$, where the
\texttt{think} field is a function-as-evidence reasoning chain written from
the perspective of directly observing the 3D geometry.  Each triplet is
branched into a \emph{direct-answer} variant (answer only, for Stage~1) and
a \emph{chain-of-thought} variant (\texttt{think} wrapped in reasoning tags,
for Stage~2).  Stage~3 starts from the Stage~2 checkpoint and trains on
re-annotated object descriptions.  The pipeline comprises three steps.

\paragraph{Step~1: Part Tree Preprocessing.}
Raw PartNeXt hierarchies are normalized by collapsing single-child chains,
merging over-branching nodes, and capping total node count, yielding compact
trees that preserve hierarchical semantics while bounding downstream MLLM
calls.

\paragraph{Step~2: Hierarchical Caption Generation.}
We traverse the tree in BFS order, captioning every node via multi-view
Set-of-Mark (SoM) rendering.  For sibling groups, six viewpoints pair a
clean GLB render (appearance) with a SoM point-cloud render (part
boundaries), and a vision-language model jointly produces a refined name
and geometry-function caption per sibling.  BFS order ensures parent
captions are always available as context.

\paragraph{Step~3: Teacher-Student CoT QA Construction.}
The annotated tree is serialized and fed to a teacher MLLM with access to
multi-view renders and ground-truth identity.  The teacher generates QA
pairs whose \texttt{think} fields are written from the student's
perspective---reasoning from geometric observations to functional inferences
to identity---producing the function-as-evidence direction.  This
asymmetry is critical: the teacher sees privileged information to
\emph{formulate} correct answers, but the student's reasoning must rely
solely on 3D geometry, mirroring the inference path at test time.
Four QA categories are generated: object identity, structural-functional
understanding, part counting, and part identity.

Applied to 23{,}138 PartNeXt objects, the pipeline yields
\textbf{507K CoT} and \textbf{249K direct-answer} QA pairs automatically.
The chain-of-thought (CoT) variant spans four QA types: part identity (283K, 55.8\%),
structural-functional understanding (129K, 25.4\%), part counting
(68K, 13.3\%), and object identity (28K, 5.5\%).
PartNeXt annotations are used only to construct these training examples;
the trained model receives raw point clouds at inference time and no part
labels are provided for ModelNet40.

\subsection{Stage~1: Multi-Source Alignment}
\label{sec:stage1}

The first stage trains a lightweight projection layer to align 3D point
cloud features with the language model's embedding space while establishing
part-aware perceptual foundations.  The training data comprises three
sources with distinct roles:

\begin{enumerate}
  \item \textbf{Re-annotated Objaverse descriptions.}  Cap3D-derived
    annotations are re-annotated with a state-of-the-art VLM for improved
    accuracy, using unmodified point clouds (8192 pts, XYZ+RGB).

  \item \textbf{Part localization QA.}  PointLLM-V2's coordinate-grounded
    part queries, providing mainly indirect part-coordinate supervision and
    spatial grounding.

  \item \textbf{Part-level functional direct-answer QA.}  Direct-answer
    pipeline output paired with per-part color-masked point clouds,
    providing direct part-functional supervision.
\end{enumerate}

These sources serve distinct roles: re-annotated descriptions maintain
general 3D comprehension, part localization supplies indirect spatial
grounding, and direct-answer QA adds explicit part-functional supervision.
Stages~2--3 subsequently convert this perceptual foundation into reasoning
gains.

Let $\mathcal{D}_\text{align} = \mathcal{D}_\text{re\text{-}ann} \cup
\mathcal{D}_\text{part\text{-}loc} \cup \mathcal{D}_\text{PF}$ denote the
union of the three sources.  Only the projector $\phi$ is trained; the 3D
encoder $g$ and the LLM $p_\theta$ are frozen.  The objective is standard
next-token prediction:
\begin{equation}
  \mathcal{L}_\text{align}(\phi) =
    -\mathbb{E}_{(P,\,q,\,a)\sim \mathcal{D}_\text{align}}
    \sum_{t=1}^{|a|}
    \log\, p_\theta\!\bigl(a_t \mid \phi(g(P)),\, q,\, a_{<t}\bigr),
  \label{eq:align_loss}
\end{equation}
where $P$ is the input point cloud, $q$ is the question, and $a$ is the
target answer.

\subsection{Stage~2: Reasoning Cold-Start via Supervised Fine-Tuning}
\label{sec:stage2}

The second stage fine-tunes the full model (projector $\phi$ + LLM
$\theta$) using the chain-of-thought variant of the pipeline annotations.
Each training example wraps the rationale and answer in explicit reasoning
tags:

\begin{quote}\small\ttfamily
<think>\\
The top section presents a flat, circular surface with slight padding,
suggesting a seating function.  Below it, a horizontal ring structure
provides foot support\ldots\\
</think>\\
<answer>\\
Bar stool.  A seating furniture piece with a flat seat and a footrest ring.\\
</answer>
\end{quote}

Let $\mathcal{D}_\text{CoT}$ denote the chain-of-thought annotations with
target sequence $c = (\texttt{<think>},\, r,\, \texttt{</think>},\,
\texttt{<answer>},\, a,\, \texttt{</answer>})$.  The SFT objective is:
\begin{equation}
  \mathcal{L}_\text{SFT}(\phi, \theta) =
    -\mathbb{E}_{(P,\,q,\,c)\sim \mathcal{D}_\text{CoT}}
    \sum_{t=1}^{|c|}
    \log\, p_\theta\!\bigl(c_t \mid \phi(g(P)),\, q,\, c_{<t}\bigr).
  \label{eq:sft_loss}
\end{equation}

This stage teaches the format and structure of function-as-evidence
reasoning.  SFT alone, however, mainly yields \emph{format imitation}: the
model produces text resembling functional reasoning but does not reliably
apply the strategy to novel point clouds.  Its modest gain on the hardest
GCS quintile suggests that it has acquired the reasoning template, but not
yet the ability to actively discover diagnostic functional structures in
unseen geometry.

\subsection{Stage~3: Dual-Objective RL with Task Arithmetic Composition}
\label{sec:stage3}

To transition the model from format imitation to active functional
discovery, we apply Group Relative Policy Optimization
(GRPO)~\cite{deepseekr1}.  A key observation is that functional reasoning,
while powerful for disambiguating geometrically similar real-world
categories, is inherently a \emph{high-risk} strategy: on stylized or
unconventional objects---common in Objaverse's artistic 3D
creations---standard functional logic does not reliably apply, causing the
model to over-commit to incorrect identities.  Robust recognition therefore
requires the model to learn not only \emph{how to reason} but also how to
\emph{avoid over-specific commitment under ambiguity}, producing rich
geometric descriptions when functional inference is unreliable.  We operationalize this via two
complementary RL paths from the same SFT checkpoint---one that pushes bold
functional inference and one that encourages descriptive
grounding---and compose the resulting models via Task
Arithmetic~\cite{ilharco2023task}.

\paragraph{GRPO optimization.}
We adopt GRPO~\cite{deepseekr1}.  For each prompt $x$, the policy
generates $G$ candidates $\{y_1, \ldots, y_G\}$ scored by reward
$r_i$.  The policy is updated by maximizing:
\begin{equation}
  \mathcal{L}_\text{GRPO} \!=
    \mathbb{E}_{x,\{y_i\}}\!\Biggl[\,
      \frac{1}{G}\sum_{i=1}^{G}
      \min\!\bigl(\rho_i\hat{A}_i,\,
        \tilde{\rho}_i\hat{A}_i\bigr)
      - \beta_\text{KL}\,D_\text{KL}\!\bigl(\pi_\theta \!\|\! \pi_\text{ref}\bigr)
    \Biggr],
  \label{eq:grpo_loss}
\end{equation}
where $\hat{A}_i = (r_i - \mathrm{mean}(\mathbf{r})) / \mathrm{std}(\mathbf{r})$ is the group-relative advantage,
$\rho_i = \pi_\theta(y_i \mid x) / \pi_\text{ref}(y_i \mid x)$,
and $\tilde{\rho}_i = \mathrm{clip}(\rho_i,\, 1{-}\epsilon,\, 1{+}\epsilon)$.

\paragraph{Reward design.}
Both RL paths train on Stage~1's re-annotated Objaverse descriptions
(source~\emph{(i)}, object-level identity QA) and differ only in reward.
For candidate $y$ and annotation $y^*$, an LLM judge
$\mathcal{J}: (y, y^*) \to \{0,1\}$ is parameterized by a judging prompt.
Both rewards are binary, but their prompts differ in strictness.
For auditability, the judge must emit one JSON object containing the parsed
class/type, matched traits, contradiction flag, and binary label; free-form
explanations are discarded.  Every model and ablation shares the prompts,
parser, and invalid-output handling:

\emph{Functional reasoning reward} ($R_\text{FR}$).
The strict judge $\mathcal{J}_\text{strict}$ requires the conjunction of
three conditions:
\begin{equation}
\begin{split}
  R_\text{FR} ={}\;
    & \mathbb{1}\bigl[\mathrm{cls}(y) \sqsubseteq \mathrm{cls}(y^*)\bigr] \\
    & \cdot\;\mathbb{1}\bigl[\mathcal{T}(y) \cap \mathcal{T}(y^*) \neq \varnothing\bigr] \\
    & \cdot\;\mathbb{1}\bigl[\lnot\,\mathrm{contra}(y, y^*)\bigr],
\end{split}
\label{eq:reward_fr}
\end{equation}
where $\mathrm{cls}(y) \sqsubseteq \mathrm{cls}(y^*)$ requires the
predicted class to be a sub-class or synonym of the ground truth (parent
categories such as ``furniture'' are rejected),
$\mathcal{T}(\cdot)$ extracts the set of key identifying or
functional traits, and $\mathrm{contra}(\cdot,\cdot)$ detects
contradictions.  By demanding both precise naming and functional evidence,
$R_\text{FR}$ incentivizes bold function-based inferences.

\emph{Descriptive grounding reward} ($R_\text{DG}$).
The lenient judge $\mathcal{J}_\text{lenient}$, aligned with the standard
evaluation protocol~\cite{pointllmv2}, checks only general type agreement:
\begin{equation}
  R_\text{DG} =
    \mathbb{1}\bigl[\mathrm{type}(y) \approx \mathrm{type}(y^*)\bigr],
  \label{eq:reward_dg}
\end{equation}
where $\approx$ means the same general type regardless of attributes.  This
permissive criterion creates an incentive toward \emph{descriptive richness}:
a hedged, feature-rich description (e.g., ``a cylindrical container with a
handle'' for ``mug'') satisfies it more reliably than a bold but potentially
wrong identity.  This is especially valuable for stylized or unconventional
Objaverse assets, where functional reasoning may over-commit; the model can
instead report faithful perceptual features that an external evaluator can
still match.

\paragraph{Task Arithmetic composition.}
The two RL paths depart from the same SFT base $\theta_\text{base}$,
producing models $\theta_\text{FR}$ and $\theta_\text{DG}$.  We define
task vectors $\boldsymbol{\tau}_\text{FR} = \theta_\text{FR} -
\theta_\text{base}$ and $\boldsymbol{\tau}_\text{DG} = \theta_\text{DG} -
\theta_\text{base}$, and compose them via linear combination:
\begin{equation}
  \theta_\text{PFIR} = \theta_\text{base}
    + \alpha\,\boldsymbol{\tau}_\text{DG}
    + \beta\,\boldsymbol{\tau}_\text{FR},
  \label{eq:task_arithmetic}
\end{equation}
where $\alpha$ and $\beta$ control the contribution of each capability.
We choose post-hoc composition over joint training with jointly combined
rewards, instantiated as $R_\text{FR}+R_\text{DG}$ or
$R_\text{FR}\!\cdot\!R_\text{DG}$ in Table~\ref{tab:ablation_dual}, for two
reasons.  First, both
rewards are binary, so a weighted combination yields limited gradient
diversity under GRPO's group-relative advantage.  Second, the two
objectives incentivize opposing behaviors---bold identification
\emph{vs.}\ conservative description---that interfere during co-optimization
(Table~\ref{tab:ablation_dual}).  Task Arithmetic sidesteps this by letting
each path optimize its objective to the fullest, then interpolating in
weight space.  The prerequisite is approximate orthogonality of the two task
vectors; we directly verify this via cosine similarity:
$\cos(\boldsymbol{\tau}_\text{FR},\,\boldsymbol{\tau}_\text{DG}) =
0.056$,\footnote{Computed over all trainable parameters
($\sim$8B).} a measured value for this pair of trained checkpoints rather
than a theoretical constant, confirming near-orthogonality.  The merged model provides a
balanced operating point: it preserves most of the hard-case accuracy of
$R_\text{FR}$ while recovering much of the in-distribution grounding of
$R_\text{DG}$ (Table~\ref{tab:ablation_dual}).  This is capability
composition rather than strict dominance.
We select $\alpha{=}0.5$, $\beta{=}0.7$ as a balanced operating point via
grid search.  Nearby weights yield similar MN40/Q5/Objaverse results:
$(\alpha,\beta){=}(0.50,0.60)$ gives 66.2/41.3/70.6, and
  $(0.50,0.90)$ gives 66.2/40.5/71.0.

\paragraph{From imitation to discovery.}
The progression reflects a deliberate pedagogical design.
Stage~1 establishes the perceptual foundation (``seeing'' parts).
Stage~2 provides the reasoning template (``knowing'' the
geometry-to-function-to-identity format).
Stage~3 trains two complementary experts: one that boldly applies
functional reasoning ($R_\text{FR}$) and one that favors rich geometric
descriptions when specific identification is uncertain ($R_\text{DG}$).
Task Arithmetic composes these experts into a single model that combines
functional reasoning with descriptive grounding---the two capabilities
needed for robust function-guided recognition.

\begin{table*}[!t]
  \centering
  \caption{Generative zero-shot 3D classification and captioning.
    (I)/\!(C): instruction-/completion-typed prompt.
    Captioning: GPT-4o image-based scoring on 200 Objaverse objects
    following~\cite{pointllmv2}.
    ModelNet40~\cite{modelnet40} is OOD and XYZ-only for all methods.
    Best in \textbf{bold}, second best
    \underline{underlined}.\protect\footnotemark}
  \label{tab:main_gen}
  \footnotesize
  \setlength{\tabcolsep}{2.5pt}
  \begin{tabular*}{\textwidth}{@{\extracolsep{\fill}}l l l l ccc ccc ccc@{}}
    \toprule
    \multirow{2}{*}{Method} &
      \multirow{2}{*}{LLM} &
      \multirow{2}{*}{3D Enc.} &
      \multirow{2}{*}{Train Data} &
      \multicolumn{3}{c}{MN40 Cls.\ (\%)} &
      \multicolumn{3}{c}{Obj.\ Cls.\ (\%)} &
      \multicolumn{3}{c}{Obj.\ Caption (Img-Based)} \\
    \cmidrule(lr){5-7} \cmidrule(lr){8-10} \cmidrule(lr){11-13}
    & & & & (I) & (C) & Avg & (I) & (C) & Avg
    & Corr & Hall$\downarrow$ & Prec \\
    \midrule
    Point-Bind LLM~\cite{pointbindllm}
      & LLaMA-7B     & I2P-MAE    & SN
      & 48.74 & 37.60 & 43.17 & 6.00 & 4.50 & 5.25
      & 0.61 & 3.77 & 13.91 \\
    PointLLM-7B~\cite{pointllm}
      & Vicuna-7B    & Point-BERT & Obj.
      & 51.54 & 50.61 & 51.08 & 61.00 & 63.50 & 62.25
      & 2.43 & 1.87 & 56.57 \\
    PointLLM-13B~\cite{pointllm}
      & Vicuna-13B   & Point-BERT & Obj.
      & 50.65 & 50.93 & 50.79 & 64.50 & 64.00 & 64.25
      & 2.48 & 1.82 & 57.84 \\
    ShapeLLM-7B~\cite{shapellm}
      & Vicuna-7B    & ReCon++    & Obj.+GAP
      & 53.08 & 53.08 & 53.08 & 54.50 & 54.50 & 54.50
      & 1.05 & 1.36 & 53.22 \\
    ShapeLLM-13B~\cite{shapellm}
      & Vicuna-13B   & ReCon++    & Obj.+GAP
      & 52.96 & 52.96 & 52.96 & 54.00 & 54.00 & 54.00
      & 1.67 & 1.44 & 53.76 \\
    MiniGPT-3D~\cite{minigpt3d}
      & Phi-2 (2.7B) & Point-BERT & Obj.
      & 57.74 & 56.24 & 56.99 & 65.50 & 66.50 & 66.00
      & 2.37 & 1.70 & 58.23 \\
    PointLLM-V2~\cite{pointllmv2}
      & LLaMA-3.1-8B & Point-BERT & Obj.
      & \underline{60.49} & \underline{61.02} & \underline{60.76}
      & \textbf{71.00} & \underline{69.00} & \textbf{70.00}
      & \underline{3.51} & \underline{1.21} & \underline{74.40} \\
    \midrule
    \textbf{PFIR (Ours)}
      & Qwen3-VL-8B  & Point-BERT & Obj.
      & \textbf{66.69} & \textbf{66.09} & \textbf{66.39}
      & \underline{70.40} & \textbf{69.27} & \underline{69.84}
      & \textbf{4.20} & \textbf{0.96} & \textbf{81.30} \\
    \midrule
    Human & -- & -- & --
      & -- & -- & -- & -- & -- & --
      & 2.61 & 0.88 & 74.79 \\
    \bottomrule
  \end{tabular*}
\end{table*}

\section{Experiments}

\subsection{Experimental Setup}

\subsubsection{Datasets}
\paragraph{Training data.}
Although the full pipeline produces 507K CoT + 249K direct-answer QA
pairs from 23{,}138 PartNeXt-annotated objects (Section~\ref{sec:pipeline}),
we retain only the
\textbf{7{,}705 objects} ($\sim$33\%) whose identifiers lie in Objaverse's
training distribution, discarding 15{,}433 non-Objaverse objects to prevent
novel-asset leakage and preserve strict OOD integrity against ModelNet40.
After filtering, 155K CoT + 85K direct-answer pairs remain.

\textbf{Stage~1} combines 618K samples over $\sim$276K unique point clouds:
\emph{(i)}~224K re-annotated Objaverse object-description QA, preserving
general 3D understanding; \emph{(ii)}~319K PointLLM-V2 part-localization
referring QA about part coordinates and names, providing spatial grounding;
and
\emph{(iii)}~75K direct functional QA over 6{,}758 objects, using per-part
color masking for part and object identity.  Source~\emph{(iii)} comprises
$\sim$68K part- and 7K object-level examples.  Overall, 62.7\% of Stage~1 is
part-level (319K + $\sim$68K) and 37.3\% is object-level (224K + $\sim$7K).
\textbf{Stage~2}'s 74K-sample reasoning cold start combines 68K PartNeXt
part-identity entries carrying function-as-evidence CoT reasoning chains
(6{,}518 objects) with 6K
reasoning-augmented Objaverse descriptions.  \textbf{Stage~3} applies GRPO
to source~\emph{(i)} (224K object-identity QA): given only a point cloud and
recognition prompt, the model must generate its own reasoning chain and is
rewarded against the ground-truth identity.  No stage sees ModelNet40.

\paragraph{Evaluation data.}
We use a strict \emph{dual out-of-distribution} (OOD) protocol.  The primary
benchmark is the ModelNet40~\cite{modelnet40} test split (2,468 samples, 40
classes), which shifts both dataset (Objaverse $\to$ ModelNet40) and
modality (XYZRGB $\to$ XYZ-only; absent color is set uniformly black).  We
additionally evaluate 3,000 in-distribution Objaverse test samples to verify
that function-guided training does not degrade general 3D understanding.

\subsubsection{Evaluation Protocol}

\paragraph{Generative zero-shot classification.}
Following PointLLM~\cite{pointllm}, we use an instruction-typed (I) question
``What is this?'' and a completion-typed (C) prompt ``This is an object of''.
The model generates free-form text with \emph{no candidate labels}.  A
shared ChatGPT post-processing prompt maps each answer to one of the 40
ModelNet40 categories.  It must return exactly one category or ``INVALID'';
matching is case-insensitive with punctuation normalized, and invalid
outputs are wrong.  The fixed prompt and parser are blind to model identity
and training configuration.  We report I, C, and average accuracy, and use
the same protocol with fixed Objaverse labels.  After removing reasoning
tags, PFIR's visible answers average 29.0/30.9 words on
ModelNet40/Objaverse; much longer no-CoT $R_\text{FR}$ outputs (121--127
words) attain only 36.1\% on Q5.  Thus verbosity does not explain the
hard-quintile gain.  All classification results use this generative
zero-shot protocol.

\paragraph{GCS-stratified quintile analysis.}
We partition ModelNet40 into five equal GCS groups
(Section~\ref{sec:problem}), from Q1 (least ambiguous) to Q5 (most confused),
and report per-quintile accuracy to expose variation hidden by aggregates.

\paragraph{In-distribution evaluation.}
Following~\cite{pointllmv2}, Objaverse evaluation reports ChatGPT-based
open-vocabulary accuracy; GPT-4o correctness, hallucination, and precision;
and Sentence-BERT~\cite{reimers2019sbert} and SimCSE~\cite{gao2021simcse}
caption similarity.

\subsubsection{Baselines}
We compare with generative 3D MLLMs: PointLLM~\cite{pointllm} (7B/13B,
Vicuna), PointLLM-V2~\cite{pointllmv2} (our direct baseline;
LLaMA-3.1-8B, 1.8M samples with part perception), ShapeLLM~\cite{shapellm}
(7B/13B, ReCon++ encoder), MiniGPT-3D~\cite{minigpt3d} (Phi-2, 2D-prior
alignment), and Point-Bind LLM~\cite{pointbindllm}.  No baseline sees
ModelNet40 during training.

\subsubsection{Implementation Details}
Our PointLLM-V2~\cite{pointllmv2} architecture uses a Point-BERT encoder
pretrained by ULIP-2 on Objaverse~\cite{ulip2,pointbert} and replaces its
LLaMA-3.1-8B~\cite{llama3} with same-class Qwen3-VL-8B~\cite{qwen3vl}.
Inputs contain $n{=}8192$, $d{=}6$ XYZRGB points (uniform black for
ModelNet40); the projector is three GELU-activated linear layers.

\textbf{Stage~1:} freeze encoder/LLM and train the projector for 3 epochs
(LR $2{\times}10^{-3}$, effective batch 128, cosine schedule, 0.03 warmup).
\textbf{Stage~2:} fine-tune projector + LLM on CoT QA for 2 epochs (LR
$2{\times}10^{-5}$, effective batch 32).  \textbf{Stage~3:} from Stage~2,
run independent $R_\text{FR}$ and $R_\text{DG}$ GRPO on identical data
(Section~\ref{sec:stage3}), each for 1 epoch with $G{=}16$, KL 0.001, LR
$1{\times}10^{-6}$, temperature 0.9, and per-GPU batch 16 on
$8{\times}$A100-80GB.  Task Arithmetic (Eq.~\eqref{eq:task_arithmetic})
uses $\alpha{=}0.5$, $\beta{=}0.7$, selected from
$\{0.3,0.5,0.7,0.9\}^2$ to maximize the joint MN40--Objaverse objective.

\footnotetext{Dataset abbreviations:
  Obj.\ = Objaverse~\cite{objaverse},
  SN = ShapeNet~\cite{shapenet},
  GAP = GAPartNet~\cite{gapartnet}.}

\subsection{Comparison with State-of-the-Art}
\label{sec:main_results}

\subsubsection{Generative Zero-Shot Classification and Captioning}

Table~\ref{tab:main_gen} presents unified results on generative
zero-shot classification (ModelNet40 and Objaverse) and 3D object
captioning (Objaverse, GPT-4o image-based evaluation).
We make the following observations:

\paragraph{OOD accuracy.}
Under the dual OOD protocol, PFIR advances instruction-prompt ModelNet40
accuracy from 60.5\% to \textbf{66.7\%}, a gain of \textbf{6.2 points},
with no ModelNet40 supervision and despite modality shift
(XYZRGB $\to$ XYZ-only).  The completion prompt shows a similar increase,
from 61.0\% to 66.1\%, and prompt-averaged accuracy rises from 60.8\% to
66.4\%; the improvement is therefore not tied to one instruction phrasing.
PFIR also surpasses ShapeLLM-13B (53.0\%, ReCon++ encoder) and MiniGPT-3D
(57.7\%), suggesting that encoder quality or architectural efficiency alone
may not resolve the generalization bottleneck targeted by our approach.

\paragraph{In-distribution and captioning.}
On Objaverse classification, PFIR averages 69.84\%, comparable to
PointLLM-V2's 70.00\%---expected since Objaverse's stylized 3D art possesses
high per-object visual distinctiveness where pattern matching already works
well.  Thus the OOD gain costs only 0.16 points on the in-distribution
average.  Meanwhile, PFIR substantially improves captioning:
correctness \textbf{4.20} (vs.\ 3.51), hallucination \textbf{0.96}
(vs.\ 1.21), and precision \textbf{81.30\%} (vs.\ 74.40\%).  Under these
GPT-4o-based automatic metrics, PFIR scores above the human-annotated
caption baseline in both correctness
(4.20 vs.\ 2.61) and precision (81.30\% vs.\ 74.79\%), while
approaching its hallucination score (0.96 vs.\ 0.88).  These results are
consistent with part-level functional training helping the model cover more
object details with fewer hallucinations.  PFIR's principal classification
contribution is therefore \emph{out-of-distribution generalization}, not
in-distribution gains.

\subsubsection{GCS Quintile Analysis}

Table~\ref{tab:gcs_quintile} stratifies ModelNet40 accuracy by GCS
quintile.  The baseline degrades from 84.0\% (Q1) to 33.7\% (Q5)---a
50-point gap invisible to aggregate metrics.  PFIR's gain is substantially
larger on Q5 than on Q1 ($+8.7$ vs.\ $+3.4$); the middle-to-hard groups
also gain +8.5 (Q3) and +8.1 (Q4), compared with +5.5 on Q2.  The benefit
is broadly larger as prototype proximity becomes less reliable.
We define a \emph{concentration ratio} measuring how disproportionately
gains accrue to the hardest quintile:
\begin{equation}
  \mathcal{C} = \frac{\Delta_5}{\Delta_1} = \frac{+8.7}{+3.4} = 2.6,
  \label{eq:concentration}
\end{equation}
supporting the hypothesis that functional reasoning benefits the most
geometrically ambiguous samples $2.6\times$ more than the least ambiguous.

\subsection{Qualitative Analysis}
\label{sec:qualitative}

\begin{table*}[!t]
\small
\centering
\caption{\textbf{Qualitative comparisons.} The classification and caption results of models on ModelNet40 and Objaverse are presented. Correct answers and fine-grained functional descriptions are highlighted in {\color{LightGreen}\textbf{green bold}}.}
\label{tab:qualitative}

\renewcommand{\arraystretch}{1}
\resizebox*{\linewidth}{!}{
\begin{tabular}{@{}l p{0.48\linewidth} p{0.48\linewidth}}
\toprule

Samples 1, 2 &
  \begin{minipage}{\linewidth}
    \centering
    \includegraphics[width=0.14\linewidth,height=0.10\linewidth]{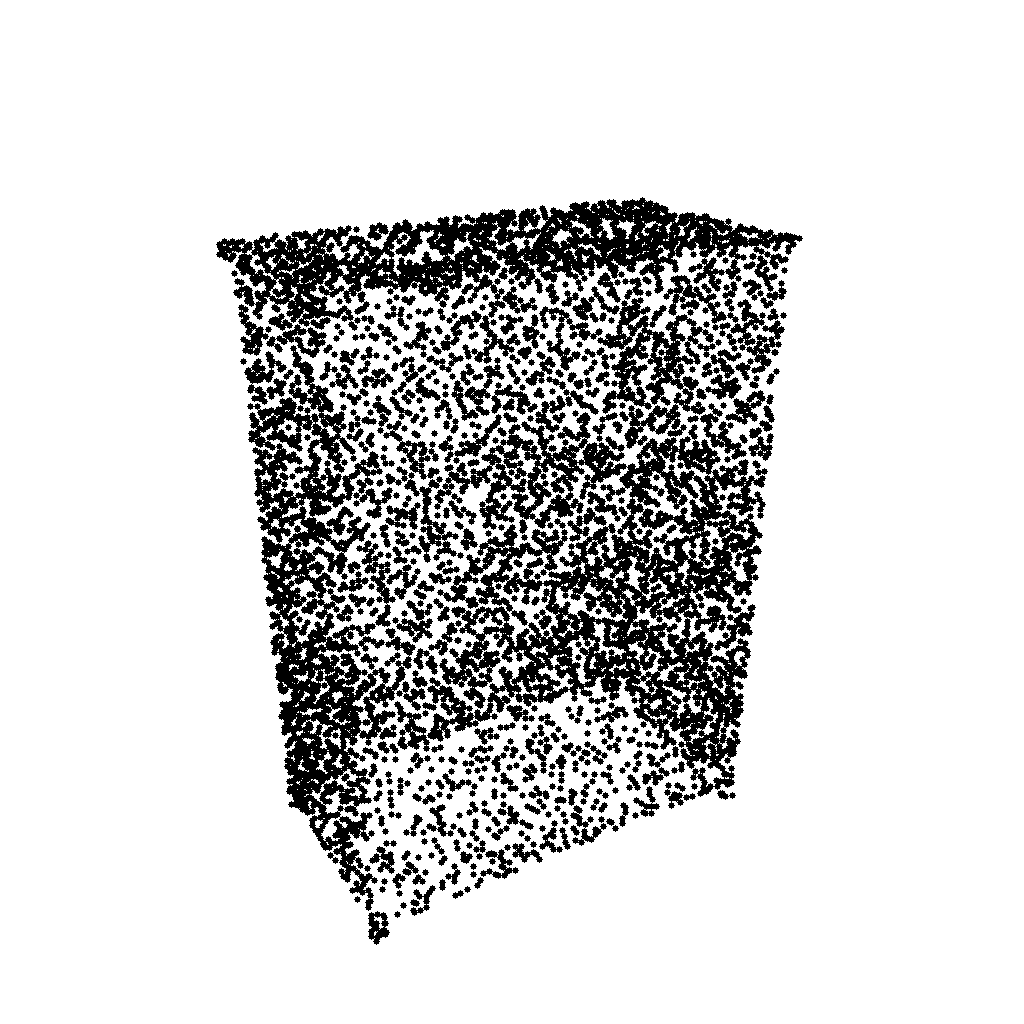}
    \hspace{0.2em}
    \includegraphics[width=0.14\linewidth,height=0.10\linewidth]{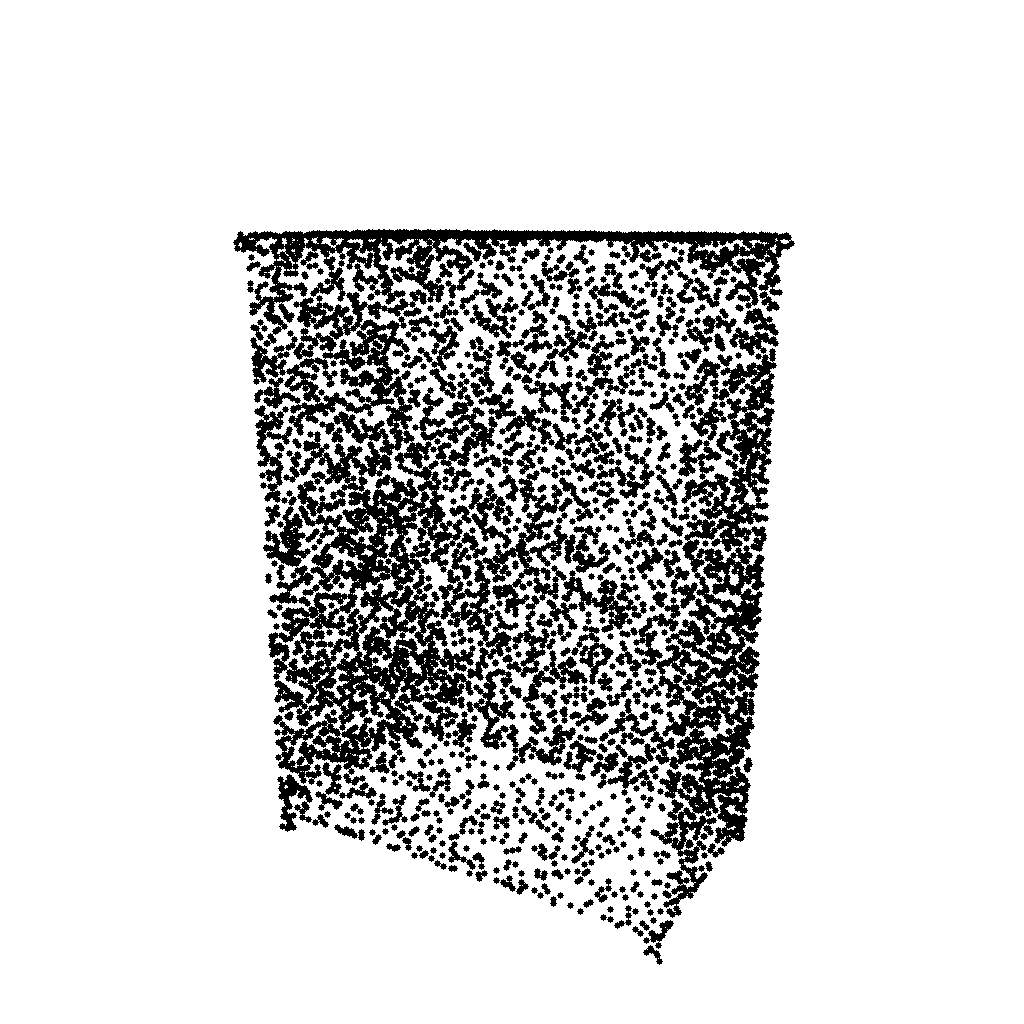}
    \hspace{0.6em}
    \includegraphics[width=0.14\linewidth,height=0.10\linewidth]{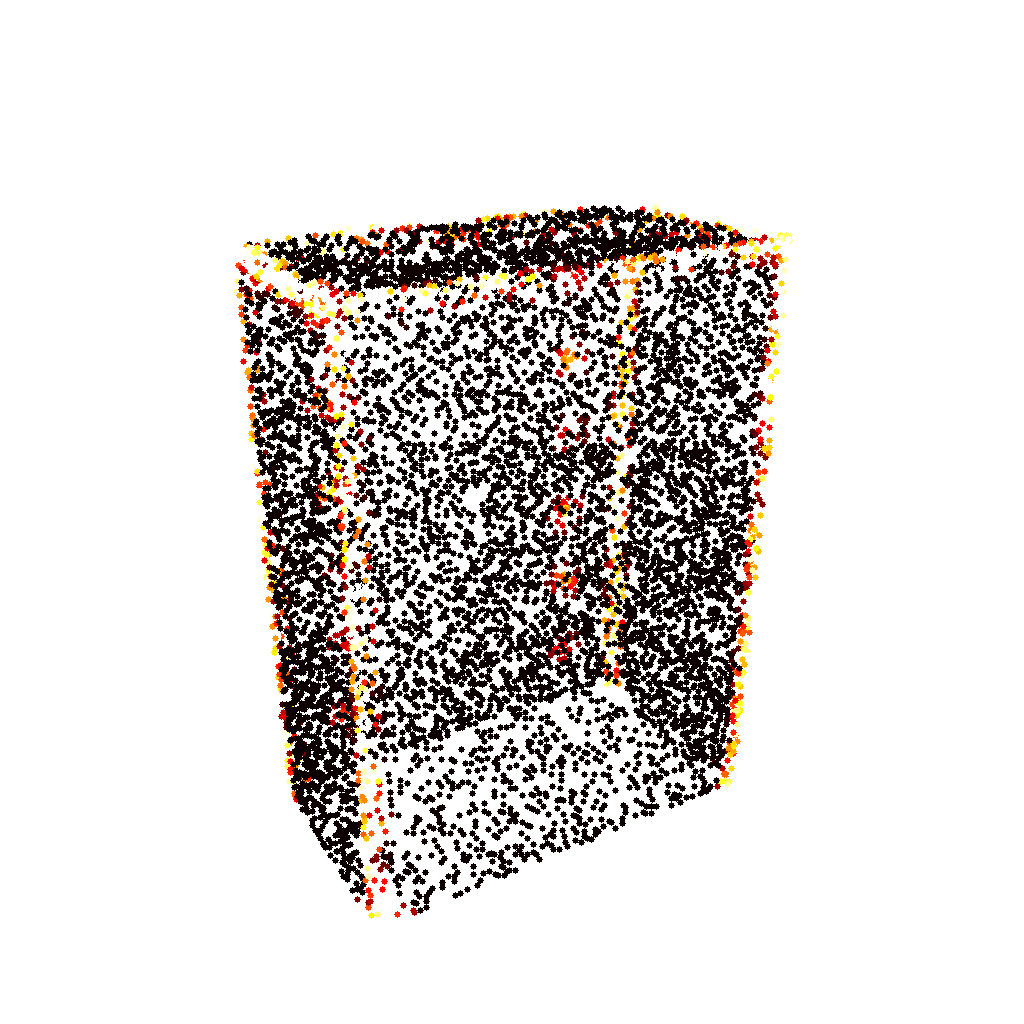}
    \hspace{0.2em}
    \includegraphics[width=0.14\linewidth,height=0.10\linewidth]{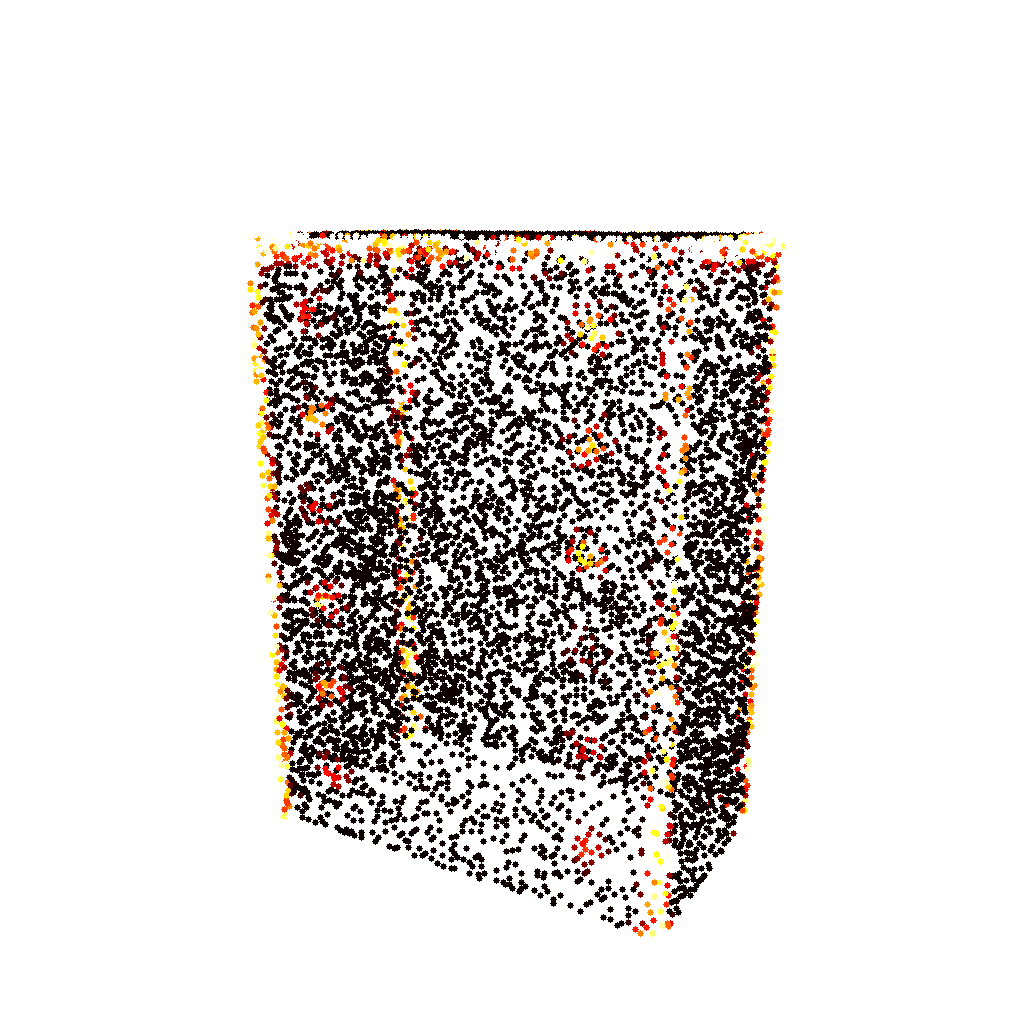}\\[-0.2em]
    {\scriptsize \hspace{0.6em} Point Cloud \hspace{3.2em} Curvature Vis.}
  \end{minipage}
&
  \begin{minipage}{\linewidth}
    \hspace{1.6em}
    \includegraphics[width=0.20\linewidth,height=0.12\linewidth]{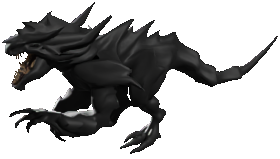}
    \hspace{1.6em}
    \includegraphics[width=0.20\linewidth,height=0.12\linewidth]{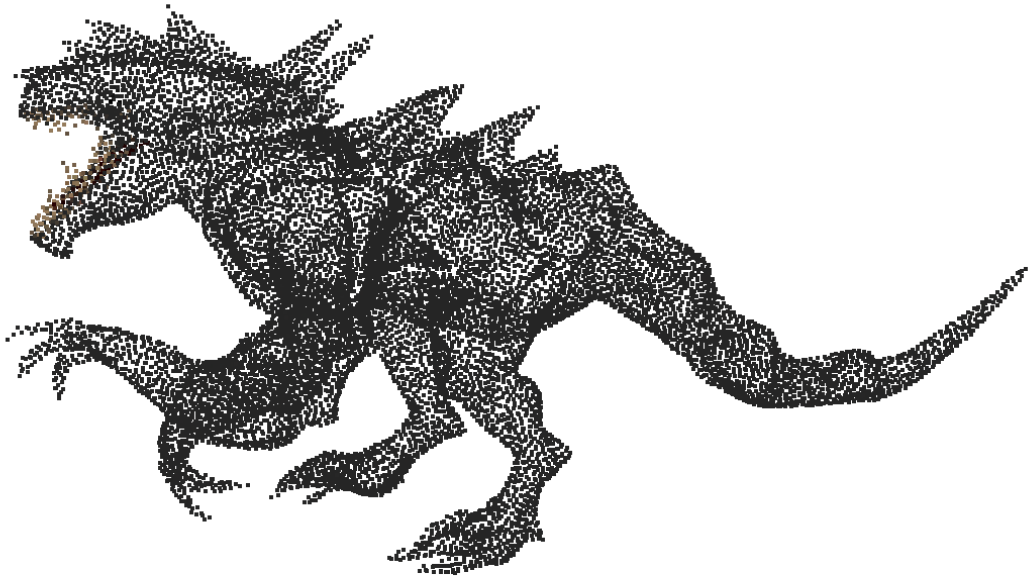}
  \end{minipage}
\\ \midrule
Prompt & What is this? & Briefly caption this 3D model. \\
Ground Truth & Dresser & A cartoon black monster like a dragon \\
\midrule

InstructBLIP-13B~\cite{dai2023instructblip} & 3d model. & a black lizard with a sharp tooth in a dark room \\
LLaVA-13B~\cite{liu2023visual} & The image features square box or cube. & A 3D model of a dark, menacing dragon. \\
3D-LLM~\cite{3dllm} & - & A black and white tiger with long legs, standing on its hind leg. \\
Point-Bind LLM~\cite{pointbindllm} & - & The 3D model features a large, ornate gargoyle with a horned helmet, sitting on top of a building. \\
PointLLM-13B~\cite{pointllm} & This is a three-dimensional model of a tall, rectangular, grey container. It stands vertically, and features a series of vents on the top. The container appears to be made of a sturdy, durable material. Its design suggests it can be used for storage, potentially for industrial purposes, due to the presumed robustness and its tall structure. Moreover, the vents on the top might be for heat dissipation, or possibly for airflow control. & The 3D model depicts a menacing black dragon, with its mouth opened wide revealing a row of sharp teeth. \\
MiniGPT-3D~\cite{minigpt3d} & This is a 3d model of a cartoon-style, black cabinet. & This is a 3D model of a fearsome, black-colored dinosaur. The dinosaur possesses an aggressive stance, with sharp claws extending from its arms and legs. It has long black fangs that seem ready to attack. \\
PointLLM-V2-8B~\cite{pointllmv2} & The object is a dark-colored, rectangular cabinet with a smooth surface. It features a series of handles on one side, indicating doors or compartments. The cabinet appears to be made of a solid material, possibly wood or metal, and has a simple, functional design without any visible decorations or embellishments. & A black dragon with spikes, horns, and a long tail. The mouth of the dragon is wide open. \\
\midrule
\textbf{PFIR (Ours)} & \texttt{<think>} This is a tall, rectangular storage unit with a dark finish. It has {\color{LightGreen}\textbf{multiple drawers}} on the front, each with {\color{LightGreen}\textbf{small round knobs}}. The top surface is \textbf{flat and slightly overhanging}, and the base is elevated by \textbf{small feet}. The overall design is simple and functional, typical of \textbf{bedroom furniture}. The structure is solid and boxy, with clean lines and no visible ornamentation. It is clearly designed for \textbf{storing clothes or other personal items}. \texttt{</think>} \texttt{<answer>} {\color{LightGreen}This point cloud represents a \textbf{dresser-style cabinet}, identifiable by its tall rectangular 3D structure and the distinct \textbf{drawer-like protrusions} on its front face.} \texttt{</answer>} & \texttt{<think>} This creature has a menacing, {\color{LightGreen}\textbf{quadrupedal form}} with a long, {\color{LightGreen}\textbf{spiky tail}} and a head adorned with {\color{LightGreen}\textbf{sharp protrusions}}. Its limbs are muscular and end in {\color{LightGreen}\textbf{clawed feet}}, suggesting a \textbf{predatory nature}. The overall silhouette is aggressive and stylized, resembling a \textbf{fictional monster} rather than a real animal. The body is covered in angular, spiky armor-like features, enhancing its fearsome appearance. Its posture is dynamic, implying movement and threat. The creature's design prioritizes intimidation and fantasy over realism. \texttt{</think>} \texttt{<answer>} {\color{LightGreen}The point cloud represents a stylized \textbf{dragon-like creature} figure, identifiable by its \textbf{spiky dorsal and tail features}, \textbf{clawed limbs}, and \textbf{elongated head structure}.} \texttt{</answer>} \\

\bottomrule
\end{tabular}
}
\end{table*}

Table~\ref{tab:qualitative} presents qualitative classification and
captioning examples on ModelNet40 and Objaverse.  Two patterns emerge:

\paragraph{Function-guided disambiguation.}
For the dresser, drawer-like protrusions and round knobs imply drawer-based
storage, supporting PFIR's identification; the baseline remains at a generic
cabinet description.  Compared with the relatively easy-to-identify wardrobe,
the dresser is a harder case that requires this part-level functional cue.
This geometry-to-function-to-identity rationale is
qualitatively consistent with function-as-evidence rather than appending
function after naming the object.

\paragraph{Graceful degradation on unambiguous samples.}
On shape-distinctive categories (e.g., dragon, airplane, car), PFIR's reasoning
remains sensible at the part and function levels
but converges to the same correct answer as the baseline, suggesting
that functional reasoning does not interfere with already-effective
shape matching.

\subsection{Ablation Study}
\label{sec:ablation}

We conduct ablation studies to examine PFIR's training design.
All ablations follow the same evaluation protocols as the main experiments.

\subsubsection{Dual-Objective RL and Composition}

Table~\ref{tab:ablation_dual} compares the two single-path RL models,
two joint-reward controls, and their Task Arithmetic composition.

\begin{table}[H]
  \centering
  \caption{GCS-aware ablation of RL strategies (\%).  PLLM-V2 denotes
    PointLLM-V2~\cite{pointllmv2}; $\dagger$ marks our reproduction.
    S1/S2 are checkpoints; S1 $\to$ RL skips Stage~2 CoT SFT.
    $\oplus$ denotes Task Arithmetic, OB-I is Objaverse instruction accuracy,
    and $\Delta$ is PFIR minus PLLM-V2.  Best/second-best scores are
    bold/underlined; bold $\Delta$ marks the largest quintile gain.}
  \label{tab:gcs_quintile}
  \label{tab:ablation_dual}
  \small
  \begin{tabular*}{\linewidth}{@{\extracolsep{\fill}}l@{\hspace{4pt}}l@{\hspace{4pt}}c c c c c c c@{}}
    \toprule
    \multirow{2}{*}{Path} & \multirow{2}{*}{Obj.} &
      \multirow{2}{*}{MN40} &
      \multicolumn{5}{c}{GCS quintile (easy $\rightarrow$ hard)} &
      \multirow{2}{*}{OB-I} \\
    \cmidrule(lr){4-8}
    & & & Q1 & Q2 & Q3 & Q4 & Q5 & \\
    \midrule
    S2 start & -- & 61.6 & 81.9 & 73.8 & 66.7 & 51.5 & 34.3 & 55.0 \\
    \cmidrule(lr){1-9}
    S1 $\to$ RL & $R_\text{FR}$ & 65.5 & \underline{88.0} & \textbf{79.7} & \underline{70.8} & 52.9 & 36.1 & 65.4 \\
    S2 $\to$ RL & $R_\text{FR}$ & \textbf{68.2} & \textbf{88.4} & \textbf{79.7} & \textbf{71.6} & \textbf{57.4} & \textbf{44.2} & 65.5 \\
    S2 $\to$ RL & $R_\text{DG}$ & 52.4 & 83.0 & 63.1 & 47.7 & 41.4 & 27.0 & \textbf{79.2} \\
    \cmidrule(lr){1-9}
    S2 joint & $R_\text{FR}\!\cdot\!R_\text{DG}$ & 65.4 & 86.4 & 76.1 & 68.2 & \underline{55.8} & 40.7 & 63.0 \\
    S2 joint & $R_\text{FR}\!+\!R_\text{DG}$ & 65.6 & \underline{88.0} & 77.5 & 69.8 & 52.1 & 39.5 & 64.4 \\
    \midrule
    \textbf{PFIR} & $\tau_\text{FR}\!\oplus\!\tau_\text{DG}$ & \underline{66.7} & 87.4 & \underline{77.7} & 70.4 & \underline{55.8} & \underline{42.3} & \underline{70.4} \\
    PLLM-V2$^\dagger$ & -- & 59.9 & 84.0 & 72.2 & 61.9 & 47.7 & 33.7 & 71.0 \\
    \textbf{$\Delta$} & -- & +6.8 & +3.4 & +5.5 & +8.5 & +8.1 & \textbf{+8.7} & $-0.6$ \\
    \bottomrule
  \end{tabular*}
\end{table}
$R_\text{FR}$ favors MN40/Q5 recognition (68.2/44.2), whereas
$R_\text{DG}$ favors OB-I grounding (79.2).  Both joint-reward controls
yield a worse MN40/Q5/OB-I balance than Task Arithmetic
(66.7/42.3/70.4), indicating interference between hard-case recognition
and in-distribution grounding.

Applying $R_\text{FR}$ after S1 without CoT reaches 65.5/36.1 on MN40/Q5
versus 68.2/44.2 after S2, isolating the cold start's value on hard cases.
PFIR retains most Q5 accuracy (42.3) while restoring OB-I to 70.4.  These
controls suggest distinct roles: SFT supplies a reusable reasoning template,
RL activates it on novel geometry, and Task Arithmetic recovers complementary
grounding.

\subsubsection{Stage-wise Contribution}

Table~\ref{tab:ablation_stage} shows the cumulative effect of each
training stage.

\begin{table}[!b]
  \centering
  \caption{Ablation on training stages.  Each row adds one stage to the
    previous configuration.  ModelNet40 instruction-prompt accuracy (\%) is
    reported.}
  \label{tab:ablation_stage}
  \small
  \begin{tabular}{l c c c c}
    \toprule
    Configuration & Stage 1 & Stage 2 & Stage 3 & MN40-I \\
    \midrule
    PointLLM-V2 baseline & -- & -- & -- & 60.5 \\
    + Multi-source alignment & \checkmark & -- & -- & 59.9 \\
    + CoT SFT & \checkmark & \checkmark & -- & 61.6 \\
    + Dual-obj.\ RL + Merge & \checkmark & \checkmark & \checkmark & \textbf{66.7} \\
    \bottomrule
  \end{tabular}
\end{table}

Stage~1 reaches 59.9\%, below PointLLM-V2's 60.5\%, because only the
projector is trained here whereas the baseline has full-model SFT; the Qwen
backbone/interface swap alone therefore gives no advantage.  Stage~2 reaches
61.6\% but only 34.3\% on Q5, consistent with format imitation.  Stage~3
gives the largest jump to \textbf{66.7\%} (+5.1\%); the merged model reaches
42.3\% on Q5, consistent with improved functional inference on difficult
samples.

\subsubsection{Data Pipeline Components}

Table~\ref{tab:ablation_data} examines the contribution of each data
source in Stage~1.

\begin{table}[!b]
  \centering
  \caption{Ablation on Stage~1 data sources.  MN40-I: ModelNet40
    instruction-prompt accuracy (\%) evaluated after Stage~1 alignment (projector-only
    training), before SFT and RL stages.}
  \label{tab:ablation_data}
  \small
  \begin{tabular}{l c c c}
    \toprule
    Re-annot. & Part-loc. & PF direct & MN40-I \\
    \midrule
    \checkmark & -- & -- & 47.4 \\
    \checkmark & \checkmark & -- & 49.5 \\
    \checkmark & \checkmark & \checkmark & \textbf{59.9} \\
    \bottomrule
  \end{tabular}
\end{table}

Table~\ref{tab:ablation_data} is a cumulative addition study.  Adding part-localization improves
Stage~1 accuracy from 47.4\% to 49.5\%, while further adding part-level
functional direct-answer QA raises it to 59.9\% (+10.4 points).  The table
therefore supports the incremental value of functional QA under the
Stage~1 protocol.

\section{Conclusion}

Our results suggest that how functional information is organized may
contribute to current 3D MLLMs' generalization bottleneck.  The per-sample
Geometric Confusion Score exposes a 50-point gap between the least and most
geometrically ambiguous samples, hidden by aggregate metrics.  This motivates
examining the prevailing \emph{function-as-annotation} paradigm, where
functional language decorates a known identity rather than providing
inference evidence.

PFIR reverses this direction through a part-level functional pipeline yielding
507K geometry $\to$ function $\to$ identity reasoning chains and a three-stage
curriculum culminating in dual-objective GRPO and Task Arithmetic, moving from
format imitation to active functional discovery.  Under strict dual OOD with
no ModelNet40 training samples, PFIR raises ModelNet40 generative zero-shot
accuracy from 59.9\% to 66.7\% in the GCS-stratified analysis; gains concentrate
$2.6\times$ more on the most ambiguous quintile, supporting part-level
functional reasoning for more generalizable 3D recognition.

\paragraph{Limitations and future work.}
PFIR requires no part annotations at inference, but its training data relies
on curated PartNeXt segmentations; robustness to predicted or noisy parts
remains untested.  Extending single-object evaluation to multi-object scenes
is a promising direction for future work.

\clearpage
\begin{acks}
This work is supported in part by the
\grantsponsor{GS501100001809}{National Natural Science Foundation of China}{https://doi.org/10.13039/501100001809}
under Grants \grantnum{GS501100001809}{62302309} and
\grantnum{GS501100001809}{62571298}.
\end{acks}

\bibliographystyle{ACM-Reference-Format}
\balance
\bibliography{references}

\end{document}